\documentclass{article}

\usepackage{arxiv}

\usepackage[utf8]{inputenc} 
\usepackage[T1]{fontenc}    
\usepackage{hyperref}       
\usepackage{url}            
\usepackage{booktabs}       
\usepackage{amsfonts}       
\usepackage{nicefrac}       
\usepackage{microtype}      
\usepackage{lipsum}
\usepackage{graphicx}
\usepackage{subfigure}
\usepackage{booktabs}
\usepackage{amsmath}
\usepackage{amssymb}
\usepackage{mathtools}
\usepackage{amsthm}
\usepackage{diagbox}
\usepackage{authblk}
\usepackage{subfigure}
\usepackage{wrapfig}
\usepackage{amsthm}
\usepackage{array}
\usepackage{amssymb}
\usepackage{bm}
\usepackage{multirow}
\usepackage{multicol}
\usepackage[capitalize,noabbrev]{cleveref}
\graphicspath{ {./images/} }

\title{Coach-assisted Multi-Agent Reinforcement Learning Framework for Unexpected Crashed Agents}

\author[1]{Jian Zhao}%
\author[1]{Youpeng Zhao}%
\author[2]{Weixun Wang}%
\author[1]{Mingyu Yang}%
\author[1]{Xunhan Hu}%
\author[1]{Wengang Zhou}%
\author[2]{Jianye Hao}%
\author[1]{Houqiang Li}%
\affil[1]{University of Science and Technology of China}
\affil[2]{College of Intelligence and Computing, Tianjin University}

\begin{document}
\maketitle
\begin{abstract}
Multi-agent reinforcement learning is difficult to be applied in practice, which is partially due to the gap between the simulated and real-world scenarios.
One reason for the gap is that the simulated systems always assume that the agents can work normally all the time, while in practice, one or more agents may unexpectedly ``crash" during the coordination process due to inevitable hardware or software failures.
Such crashes will destroy the cooperation among agents, leading to performance degradation.
In this work, we present a formal formulation of a cooperative multi-agent reinforcement learning system with unexpected crashes.
To enhance the robustness of the system to crashes, we propose a coach-assisted multi-agent reinforcement learning framework, which introduces a virtual coach agent to adjust the crash rate during training.
We design three coaching strategies and the re-sampling strategy for our coach agent.
To the best of our knowledge, this work is the first to study the unexpected crashes in the multi-agent system.
Extensive experiments on grid-world and StarCraft II micromanagement tasks demonstrate the efficacy of adaptive strategy compared with the fixed crash rate strategy and curriculum learning strategy.
The ablation study further illustrates the effectiveness of our re-sampling strategy.
\end{abstract}


\section{Introduction}
\label{chap:intro}
Cooperative multi-agent systems widely exist in various domains, where a group of agents need to coordinate with each other to maximize the team reward \cite{busoniu2008comprehensive,tuyls2012multiagent}. Such a setting can be broadly applied in the control and operation of robots, unmanned vehicles, mobile sensor networks, and the smart grid~\cite{zhang2021decentralized}.
Recently, many researchers have devoted their efforts to leveraging reinforcement learning techniques to multi-agent systems~\cite{sunehag2017value,rashid2018qmix,wang2020qplex, wang2020cooperative}.
Despite the remarkable advance in academia, multi-agent reinforcement learning is still difficult to be applied in practice.
One non-trivial reason is that there always exists a gap between the simulated and real-world scenarios, which degrades the performance of the policies once the models are transferred into real-world applications~\cite{zhao2020sim}.

To close this sim-to-real gap and accomplish more efficient policy transfer, multiple research efforts are therefore now being directed towards identifying the causes for such gap and proposing corresponding solutions.
One main cause is the difference between the physics engine of the simulator and the real-world scenario.
To alleviate the difference, research efforts have been directed towards building up more realistic simulators by mathematical models~\cite{shah2018airsim,dosovitskiy2017carla,furrer2016rotors,mccord2019distributed,todorov2012mujoco,wang2021pre}.
Another cause is the mismatch of the data distribution of the simulation environment and the real environment, which has inspired related research on domain adaptation~\cite{higgins2017darla,traore2019continual,arndt2020meta}, and domain randomization~\cite{tobin2017domain}.

Generally, the simulated systems always assume that the agents can work normally all the time.
However, this assumption is usually not in line with reality.
Due to inevitable hardware or software failures in practice, one or more agents may unexpectedly ``crash" during the coordination process.
If the agents are trained in an environment without crashes, they only master how to cooperate in a crash-free environment.
Once some agents ``break down" and take abnormal actions, the remaining agents can hardly maintain effective cooperation, which will lead to performance degradation.
Take a two-agent system as an example:
the two agents are required to finish two tasks in coordination;
under the crash-free scenario, the optimal solution is that each agent takes responsibility for one task, respectively;
when applying such a policy to the real-world application, the cooperation cannot be accomplished if any agent encounters a crash.
This example indicates the necessity of considering unexpected crashes during training in order to obtain well-trained agents with high robustness.

To the best of our knowledge, this work is the first to study the crashes in the multi-agent systems, which is more consistent with real-world scenarios.
In this study, we give a formal formulation of a cooperative multi-agent reinforcement learning system with unexpected crashes, where any agent has a certain probability of crashing during operation.
We assume that, for each agent, the probability of crashing independently follows a Bernoulli distribution.
To enhance the robustness of the system to unexpected crashes, the agents should be trained in an environment with crashes.
The key challenge is how to adjust the crash rate during training.

In this work, we propose a coach-assisted multi-agent reinforcement learning framework, which introduces a virtual coach agent into the system.
The coach agent is responsible to adjust the crash rate during training.
One straightforward coaching strategy for ``coach" is to set a fixed crash rate during training.
Considering that it may be too difficult for agents to cooperate from scratch~\cite{narvekar2020curriculum}, increasing the crash rate gradually is another feasible strategy.
In addition to these basic strategies, an experienced ``coach" can also automatically adjust the crash rate corresponding to the overall performance during training.
Specifically, if the performance exceeds the threshold, the crash rate is increased to increase the difficulty of learning; otherwise, the crash rate should be decreased.
In this way, the agents can learn the coordination skills progressively in face of the unexpected crashes.

To test the effectiveness of our method, we conduct experiments on grid-world and StarCraft II micromanagement tasks.
Compared to the fixed crash rate and curriculum Learning strategies, the results demonstrate that adaptive method achieves relatively stable performances in the case of different crash rates.
Furthermore, the ablation study shows the efficacy of our re-sampling strategy.

\section{Related Work}
In this section, we briefly summarize the works related to cooperative multi-agent reinforcement learning (MARL).
With the development of this field, researchers pay more and more attention to the MARL problem that is more consistent with real-world settings.

Early efforts treat the agents in a team independently and regard the team reward as the individual reward~\cite{tan1993multi,mnih2015human,omidshafiei2017deep,foerster2017stabilising}.
In this way, the MARL task is transformed into multiple single-agent reinforcement learning tasks.
While trivially providing a possible solution, these approaches pay insufficient attention to an essential characteristic of MARL---coordination among agents. In other words, it will bring non-stationarity that agents cannot distinguish between the stochasticity of the environment and the exploitative behaviors of other co-learners \cite{lowe2017multi}.

Another line of research focuses on centralized learning of joint actions, which can naturally take coordination problems into consideration~\cite{sukhbaatar2016learning,peng2017multiagent}.
Most of the centralized learning approaches require communication during execution.
For instance, CommNet~\cite{sukhbaatar2016learning} designs a centralized network for agents to exchange information.
BicNet~\cite{peng2017multiagent} leverages the bi-directional RNNs for information sharing.
Considering the communication constraint in practice, SchedNet \cite{kim2019learning} is proposed, in which agents learn how to schedule themselves for message passing and how to select actions based on received partial observations.
Another challenge of centralized learning is the scalability issue since the joint action space grows exponentially as the number of agents increases.
Some works investigate the scalable strategies in the centralized learning~\cite{guestrin2001multiagent,kok2006collaborative}.
Sparse cooperative Q-learning~\cite{kok2006collaborative} only allows the necessary coordination between agents by encoding such dependencies.
However, these methods require prior knowledge of the dependencies among agents, which is often inaccessible.

To study a more practical scenario with partial observability and communication constraint, an emerging stream is the paradigm of centralized training with decentralized execution (CTDE) \cite{oliehoek2008optimal,kraemer2016multi}.
To the best of our knowledge, value decomposition networks (VDN)~\cite{sunehag2017value} makes the first attempt to decompose a central state-action value function into a sum of individual Q-values to allow for decentralized execution.
VDN simply assumes the equal contributions of agents and does not make use of additional state information during training.
Based on VDN, QATTEN~\cite{yang2020qatten} utilizes a multi-head attention structure to distinguish the contributions of agents, and linearly integrates the individual Q-values into the central Q-value.
Instead of using linear monotonic value functions, QMIX~\cite{rashid2018qmix} and QTRAN~\cite{son2019qtran} employ a mixing network satisfying Individual-Global-Max (IGM) principle~\cite{son2019qtran} to combine the individual Q-values non-linearly by leveraging state information.
QPLEX~\cite{wang2020qplex} introduces the duplex dueling structure and decomposes the central Q-value into the sum of individual value functions and a non-positive advantage function.

However, all of the existing works assume that agents can continuously maintain normal operations, which is inconsistent with real-world scenarios.
As a matter of fact, it is a quite common phenomenon that some agents encounter unexpected crashes because of hardware or software failure.
To this end, our work aims to study a more practical problem by considering unexpected crashed agents in the cooperative MARL task.

\section{Problem Formulation}\label{sec:prob}
In order to better solve the problem of unexpected crashed agent, we define a Crashed Dec-POMDP model, which is defined by a tuple $M = < \bm{N}, \bm{S}, \bm{A}, \bm{\Omega}, P, O, R, \gamma, \alpha> $. 
Each agent  $g_i \in \bm{N} \equiv \{g_1, g_2,\cdots, g_n\}$ has a probability of crashing and the crash rate is denoted as $\alpha$.
For simplicity, we assume that the crash happens at the beginning of the episode and the status of being crashed or not will not change throughout the episode.
We define a binarized vector to denote the $\emph{crashed state}$ of $n$ agents as $[c_i]^n_{i=1}$, where $c_i\sim\emph{Bernoulli}(\alpha)$. 
When the $i^{th}$ agent is crashed, $c_i$ is 1; otherwise, $c_i$ is 0.
Note that $[c_i]^n_{i=0}$ stays the same during an episode but may change throughout the task due to randomness.

At each time step, each agent $g_i$ receives partial observation $o_i \in \bm{\Omega}$ according to the observation probability function $O(o_i|s)$.
Each uncrashed agent chooses an action $a_i \in \bm{A}$ with the normal strategy, while the crashed agents take ``no-move" or random actions, forming a joint action $\bm{a} =[a_i]_{i=1}^n$. 
Given the current state $s$, the joint action $\bm{a}$ of the agents transits the environment to the next state $s'\in \bm{S}$ according to the state transition function $P(s'|s, \bm{a})$. All of the agents share a team reward $R(s, \bm{a})$. 
The learning goal of MARL is to optimize every agent's individual policy $\pi_i(a_i|\tau_i)$, where $\tau_i = (o^0_i,a^0_i,o^1_i,a^1_i,\cdots,o^T_i,a^T_i)$ is an agent's action-observation history, so as to maximize the team reward accumulation $\sum_{t=0}^{\infty} \gamma^t R(s^t,\bm{a}^t)$, where $\gamma \in [0,1)$ is a discount factor.


\begin{figure*}[t]
	\centering
	\includegraphics[width=0.85\textwidth]{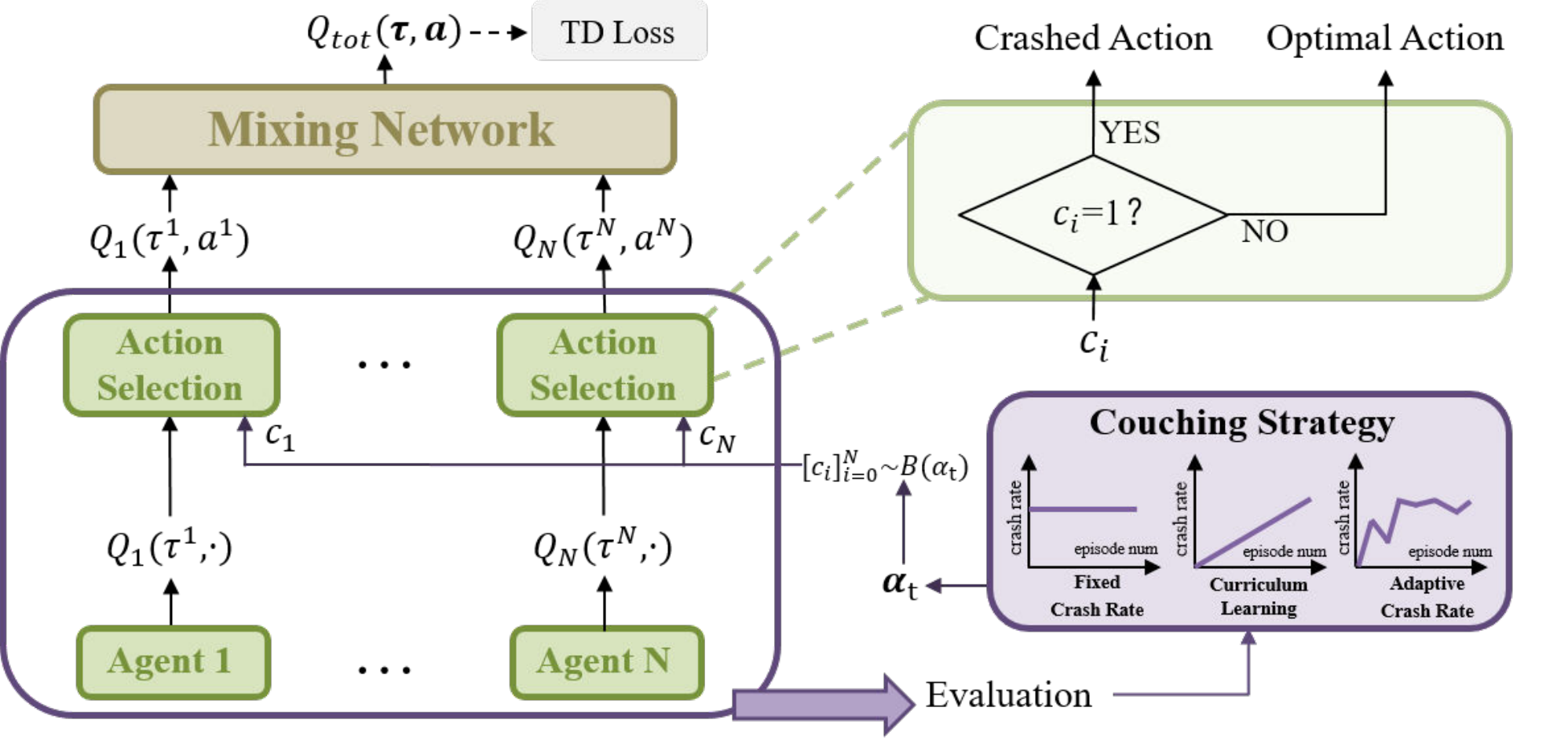}
	\caption{An overview of the adaptive framework.}
	\label{fig:model}
\end{figure*}

\section{Method}
In this section, we present our coach-assisted multi-agent reinforcement learning framework for Crashed Dec-POMDP problem and explain the rationality of our design.

\subsection{Overall Framework}
To simulate the crash scenarios during training, we introduce a virtual agent into the system to act as a ``coach".
The ``coach" is responsible for deciding the crash rate during training.
At the beginning of each episode $t$, the ``coach" sets up a crash rate $\alpha_t$.
We assume the probability of being crashed for each agent following a Bernoulli($\alpha_t$) distribution.
Given the current crash rate $\alpha_t$, some of the agents become crashed and cannot take rational actions.
Then, the multi-agent system with crashed agents are trained for $T$ steps to learn coordination.
Afterwards, the ``coach" can receive the performance of the agents under the current crash rate, denoted as $e_t$, and reset the crash rate $\alpha_{t+1}$ for the next episode.

\subsection{Coaching Strategy}
The main challenge for the ``coach" is how to choose an effective crash rate during training.
Here, we introduce three coaching strategies.

\textbf{Fixed Crash Rate.}
The ``coach" sets a fixed crash rate throughout the training process.
The agents, of which some are crashed, are required to learn coordination skills from scratch.

\textbf{Curriculum Learning.}
The ``coach" linearly increases the crash rate during training.
At the very beginning, the agents are trained under a crash-free environment.
For the $t^{th}$ episode, the ``coach" sets the crash rate to be $(t-1)\Delta\alpha$, where $\Delta\alpha$ is a hyperparameter.
In this way, the difficulty of cooperation increases gradually.

\textbf{Adaptive Crash Rate.}
For the former two strategies, the ``coach" does not take full advantage of the performance of the cooperative agents.
An advanced strategy for the ``coach" is to adaptively adjust the crash rate corresponding to the performance of the agents under the current crash rate.
The basic idea is that if the agents can cooperate well and achieve acceptable performance under the current crash situation, the crash rate should be increased; otherwise, the crash rate should be decreased.
The adaptive strategy can be formulated as follows:
\begin{equation}
    \alpha_{t+1} = F(\alpha_t, e_t, \beta),
\end{equation}
where $F(\cdot)$ is a mapping function, $e_t$ refers to the performance of the current model and $\beta$ represents the threshold of the performance of the specific evaluation metric.
We can see that the fixed crash rate and the curriculum learning are the two special cases of the adaptive strategy.

\noindent For the fixed crash rate strategy,
\begin{equation}
    F(\alpha_t, e_t, \beta) = \alpha_t.
\end{equation}
For the curriculum learning strategy,
\begin{equation}
    F(\alpha_t, e_t, \beta) = \alpha_t + \Delta\alpha,
\end{equation}
where $\alpha_1 = 0$. 

\noindent In this work, we use the following adaptive function:
\begin{equation}
    F(\alpha_t, e_t, \beta) = \alpha_t + \rho\times(I(e_t>=\beta)-\alpha_t).
\end{equation}
It is notable that our method is not limited to the selection of the above function. A more efficient adaptive function can be further investigated.

\subsection{Re-sampling Strategy}
Randomly sampling from a Bernoulli($\alpha$) distribution may cause the proportion of the crashed agents far beyond or fewer than the current crash ratio $\alpha$.
Therefore, we employ a re-sampling strategy to ensure that the number of the crashed agents is no more than the upper bound of $n\times \alpha$.
Here, we explain the rationality behind the re-sampling strategy.
For the samples with more crashed agents, it may be too difficult for the current model to learn the coordination skills, thus they are discarded.
For the samples with fewer crashed agents than the expectation, they can help the agents remember how to deal with the easier scenarios, thus they are utilized during training.

\section{Experiments}\label{sec:exp}
In this section, we conduct experiments to demonstrate the effectiveness of the methods that we propose.
We firstly conduct experiments in a grid-world environment as a toy example. Then we use StarCraft Multi-Agent Challenge (SMAC) environment ~\cite{samvelyan19smac} as the test-bed to evaluate our methods, which has become a commonly-used benchmark for evaluating state-of-the-art MARL approaches.
All experiments are conducted on a Ubuntu 18.04 sever with 4 Intel(R) Xeon(R) Gold 6252 CPU @ 2.10GHz and GeForce RTX 2080Ti GPU. Our codes are available at https://github.com/youpengzhao/Crashed\_Agent.

\subsection{Grid-world Example}
\subsubsection{Settings}
We utilize the grid-world example to intuitively show the consequence without considering unexpected crashes in real-world scenarios. We set a 10$\times$10 grid where two agents need to touch two buttons within limited steps. The game will terminate after 20 steps or both buttons have been touched. The default reward at each step is -1 and if one button is touched, the agents are assigned a reward of 5 at this step. In this way, the agents are encouraged to touch the button as quickly as possible. At each step, each agent has 5 possible actions including up, down, left, right and staying still. If there exists an unexpected crash during the test, the crashed agent will stay still through the whole episode so only one agent may crash during the test. For simplicity, the initial locations of agents and buttons are fixed so that this environment is deterministic. What's more, the observation of each agent is its own location and the global state contains the locations of the two buttons and agents so that the agents will not know whether their partner is crashed according to its own observation during operation.

We take QMIX~\cite{rashid2018qmix}, a state-of-the-art value-based MARL algorithm, as the base model in this toy example and we adopt the adaptive approach in comparison. Our implementation is based on the Pymarl algorithm library~\cite{samvelyan19smac} and training schedules such as the optimizer and training hyperparameters are kept the same as the default ones used in Pymarl. Our method includes two additional hyperparameters: one is the performance threshold $\beta$ to decide whether to increase or decrease the crash rate during training; another is the learning rate of the crash rate $\rho$ to control the stepsize of adjustment of $\alpha$. We set $\beta$ as 0.75 and $\rho$ as 0.01 in this experiment. These tasks are trained for 2 million steps separately.

\begin{figure}[htbp]
	\centering
	\subfigure[Agent1 is crashed]{
		\includegraphics[width=0.22\textwidth]{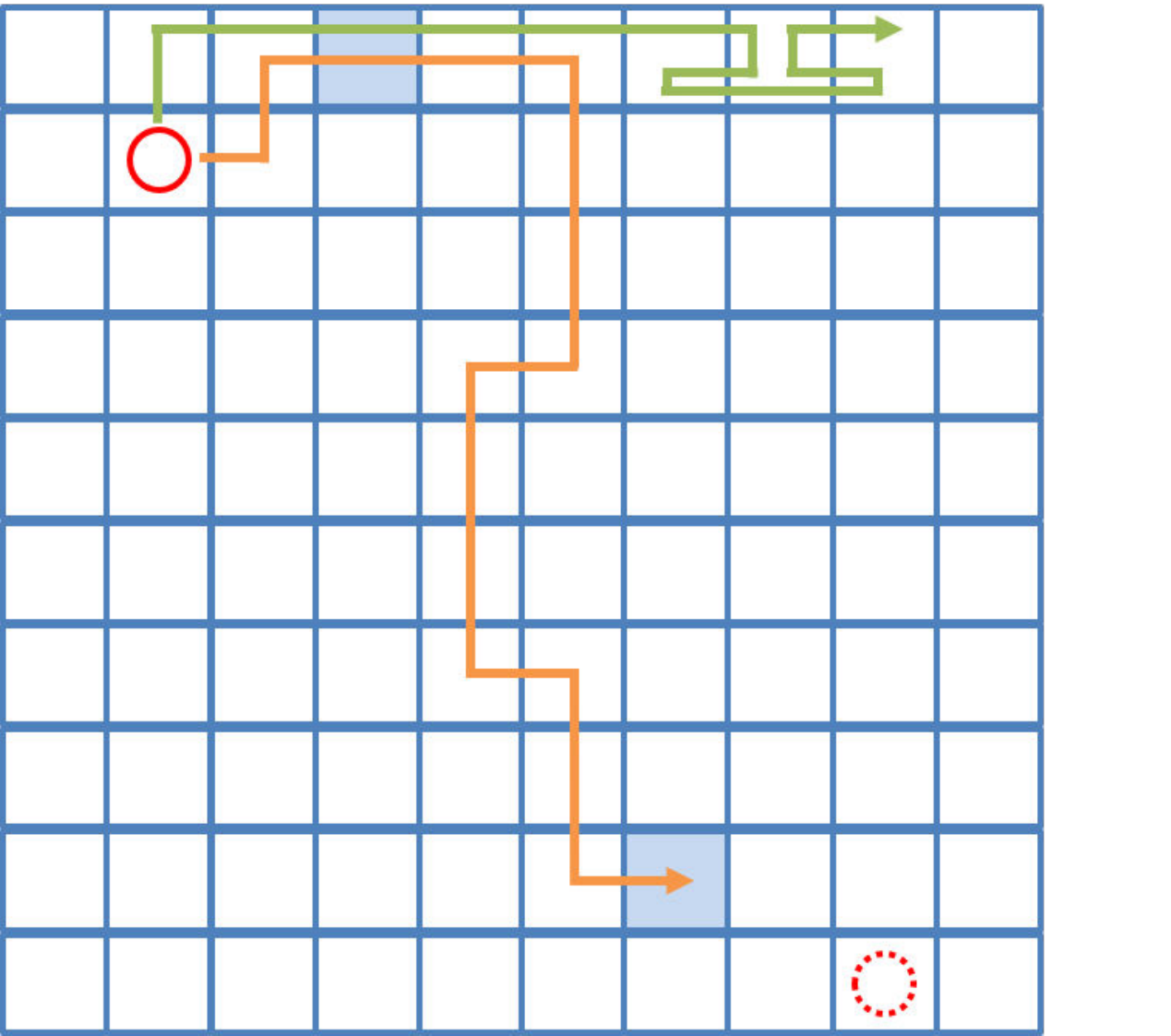}
		\label{fig:grid_1}
	}
	\subfigure[Agent2 is crashed]{
		\includegraphics[width=0.22\textwidth]{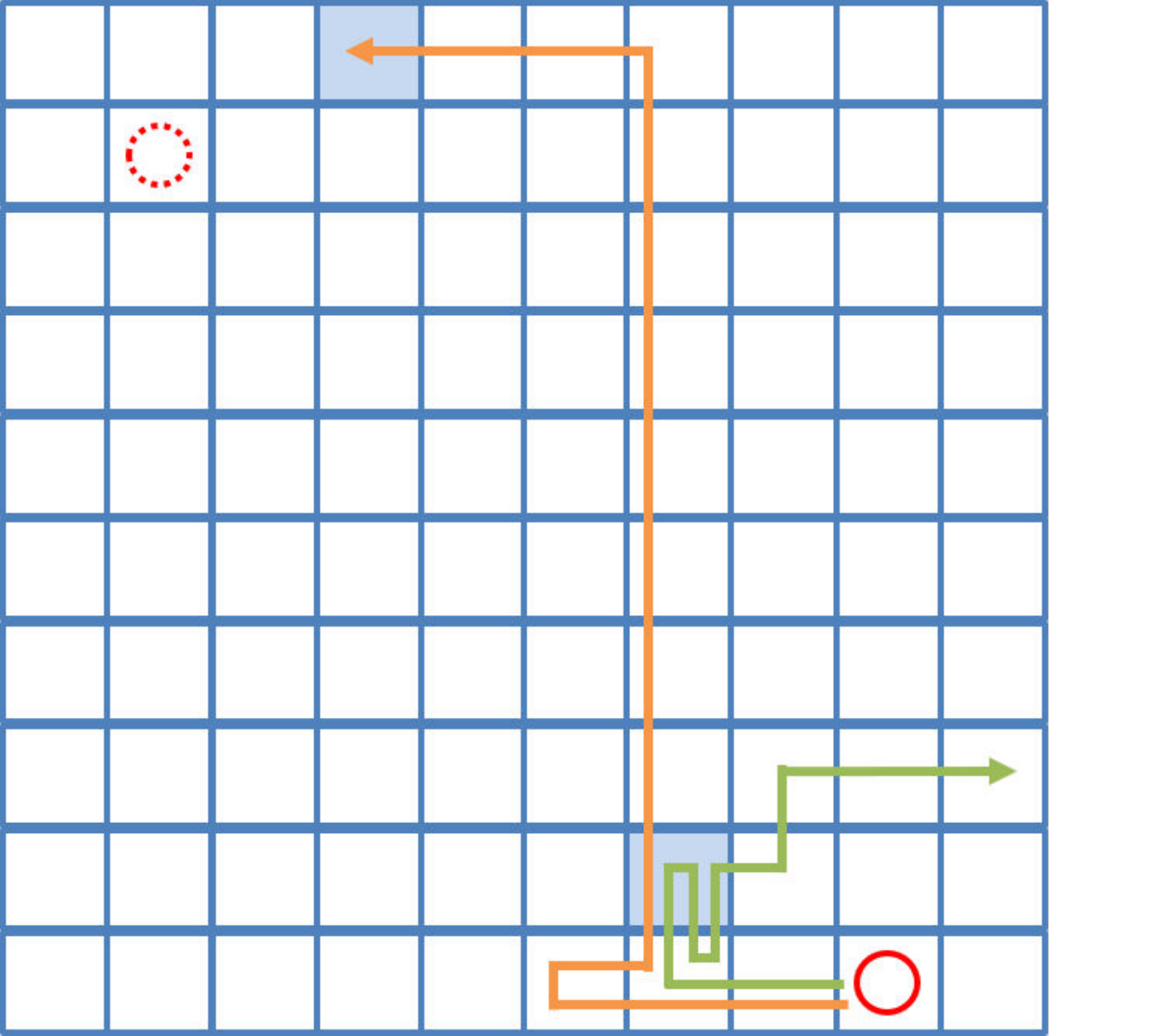}
		\label{fig:grid_2}
	}
	\caption{The trajectory of agents during test when one of them is crashed. The agents are represented with circles while the crashed one is marked using dotted line and the normal one is marked using solid line. The colored grids symbolize the two buttons. The green arrow line is the trajectory of agents trained using original QMIX and the orange one is achieved by our method. }
	\label{fig:grid}
\end{figure}

\subsubsection{Performance Evaluation and Discussion}
After training for the same steps, agents trained using both these methods manage to complete this task under normal scenarios. However, when an unexpected crash happens, things are different. The results are illustrated in Figure~\ref{fig:grid}. The agent trained using QMIX learns to touch the button near it in the shortest path but after that, it only wanders meaninglessly. Due to partial observation, the normal agent fails to know that its partner is out of control and it also does not try to touch another button. We assume that agents trained by QMIX learn to efficiently cooperate to complete this task, therefore they just need to touch the button near themselves. However, their excessive reliance on cooperation makes the system fragile and they fail to deal with the unexpected crash, which is common in realistic scenes. In contrast, our method takes possible crashes into account during training so that it can still fulfill the task even when one of them ``breaks down''. This toy example illustrates the drawback of overreliance on cooperation and the necessity of considering possible crashes when training the multi-agent system.

\begin{figure*}[!ht]
	\centering
	\subfigure[]{
		\includegraphics[width=0.44\textwidth]{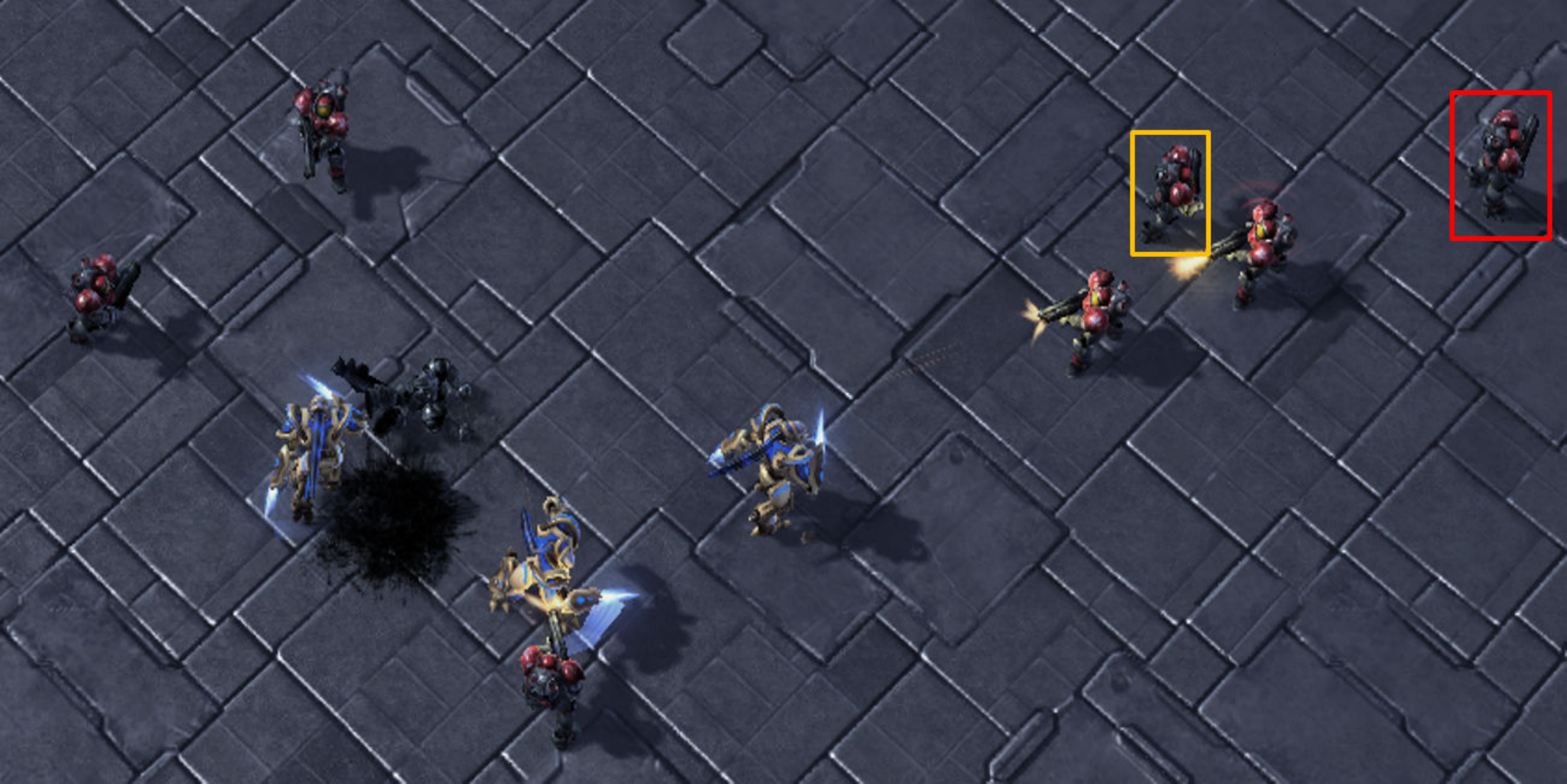}
		\label{fig:obs1_1}
	}
	\subfigure[]{
		\includegraphics[width=0.44\textwidth]{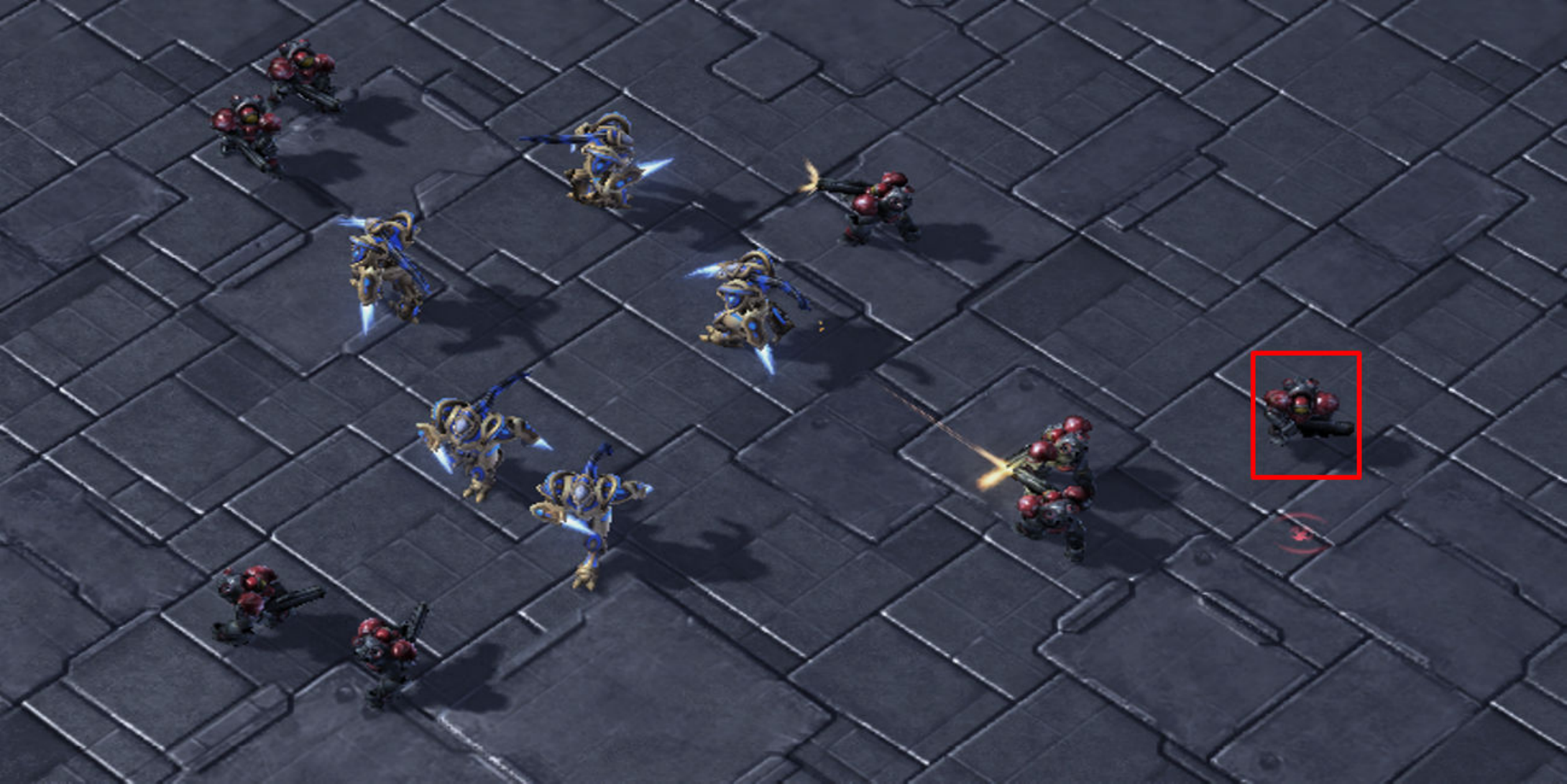}
		\label{fig:obs1_2}
	}
	\subfigure[]{
		\includegraphics[width=0.44\textwidth]{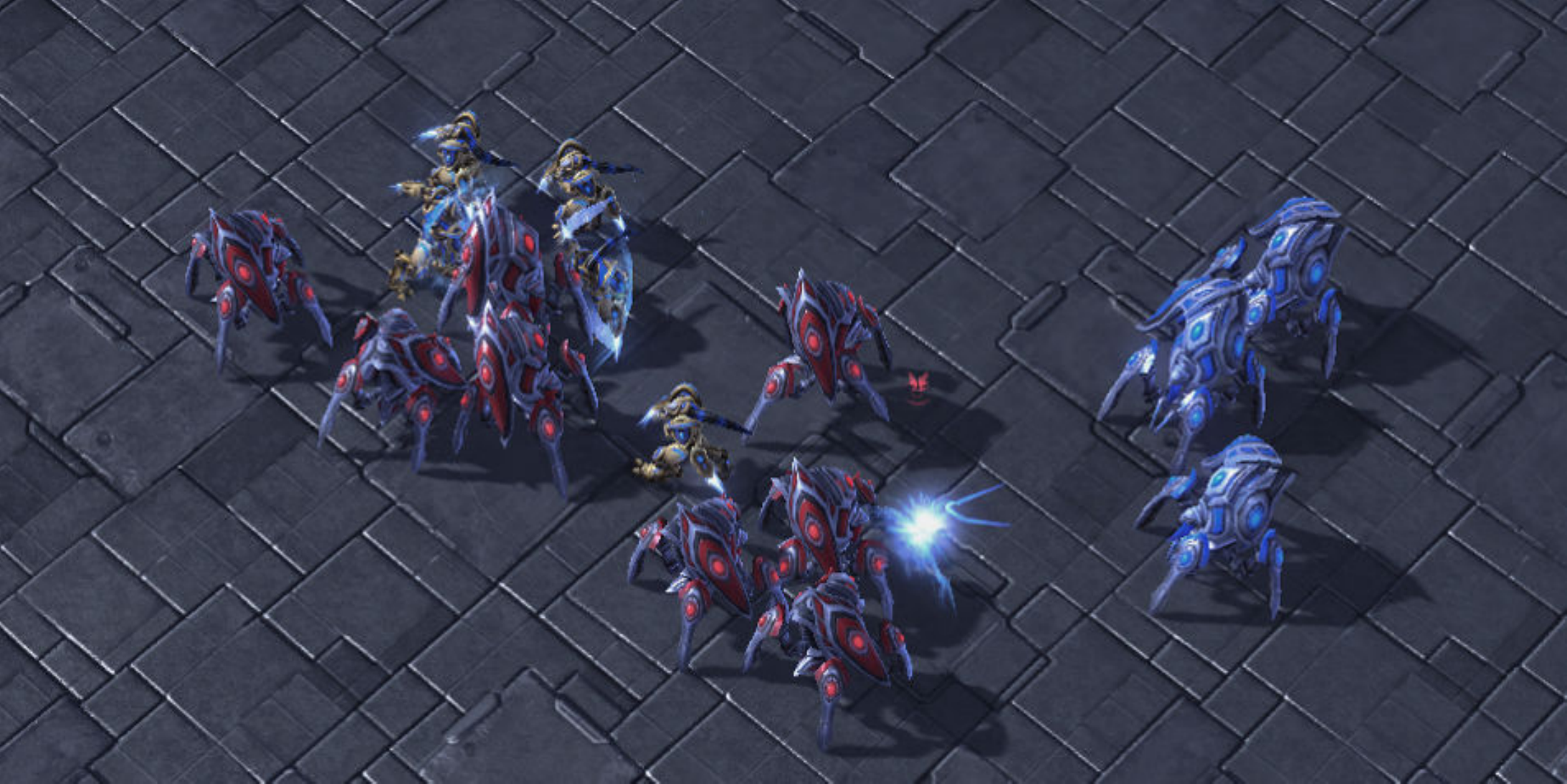}
		\label{fig:obs2_1}
	}
	\subfigure[]{
		\includegraphics[width=0.44\textwidth]{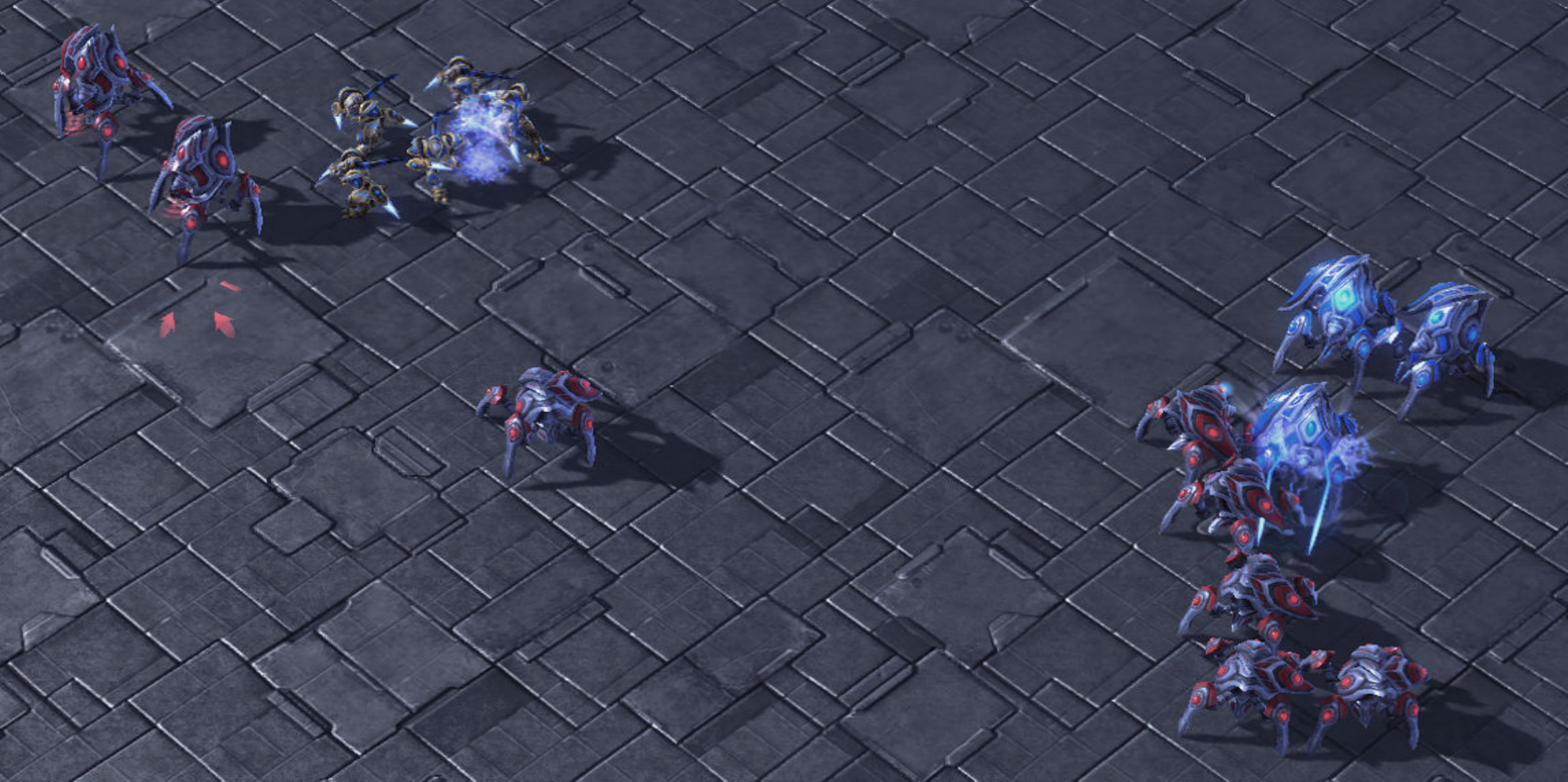}
		\label{fig:obs2_2}
	}
	\caption{The illustration of the actions taken by the agents who are trained using original QMIX (on the left) and our adaptive approach (on the right) when they are tested under scenarios with crashes. The agent highlighted with a red rectangle represents the crashed one and the agent highlighted with a yellow rectangle refers to the one that is affected.
	}
	\label{fig:ob}
\end{figure*}

\subsection{StarCraft Multi-Agent Challenge}
\subsubsection{Settings}
Apart from the experiment of grid-world, we also conduct experiments on StarCraft II decentralized micromanagement tasks to show the effectiveness of our method. In this environment, we assume the crashed agents will take random actions. Here, we also take QMIX~\cite{rashid2018qmix} as the base model. 
Then we compare the performance of QMIX and our coach-assisted framework with ``fixed crash rate'', ``curriculum learning'' and ``adaptive crash rate'' coaching strategy. Our implementation is also based on the Pymarl algorithm library~\cite{samvelyan19smac} without changing the default training schedules. 
For the variants of QMIX with the fixed crash rate, we randomly sample the crashed agents with a Bernoulli distribution during each episode, thus the actual number of the crashed agents ranges from 0 to $n$. In curriculum learning coach strategy, the crash rate increases from 0 linearly and the upper limit is set to 0.1 as we test the models in scenarios whose crash rate is at most 0.1. We set the two hyperparameters $\beta$ in \{0.60, 0.65, 0.70, 0.75\} and $\rho$ in \{0.001, 0.003, 0.005, 0.015\}, and select their optimal values based on grid search when adopting our adaptive method.
We repeat the experiments under each setting over 5 runs with different seeds and report the average results.
For all the compared methods, each task is trained for 2 million steps separately. 
In order to obtain a relatively robust evaluation result, each model is tested for 128 times.

We choose two standard maps and design two different maps in the experiment, which are 3s$\_$vs$\_$5z, 3s5z$\_$vs$\_$3s5z, 8m$\_$vs$\_$5z and 8s$\_$vs$\_$3s5z. The two standard maps are well-matched in strength so a crash may result in some imbalance. In order to comprehensively show the performance of our method, we also design two maps that guarantee an appropriate gap in strength between the two sides so that unexpected crashes will not lead to a significant change in difficulty.
To know more details about the maps, please refer to~\cite{samvelyan19smac}.

\subsubsection{Observation}
In this part, we discuss the observations from the scenario with crashed agents on StarCraft II micromanagement tasks and show what to be considered to deal with the crash scenario. 

The agents in Figure~\ref{fig:obs1_1} and \ref{fig:obs1_2} play as Marines (ours) that are good at long-range attacks while Zealots (opponents) can only attack in short-range and they have the same moving speed. Because Marines only have half health point of Zealots, the optimal strategy is to alternate fire to attract the enemies. Figure~\ref{fig:obs1_1} shows the case when one agent (highlighted with a red rectangle) is out of control and starts to take random actions, one of the remaining agents (highlighted with a yellow rectangle) is disrupted so that it cannot take reasonable action as well. This case illustrates that the random crashes of some agents will undermine the coordination among the rest of the agents in the team, which is likely to cause a drop in the win rate. However, it can be observed in Figure~\ref{fig:obs1_2} that agents trained with our method can avoid such effect of the unexpected crash as they may be familiar with abnormal observations.

Figure~\ref{fig:obs2_1} and \ref{fig:obs2_2} describe another situation where Stalkers (ours) play against Zealots as well as Stalkers (opponents) in the map 8s$\_$vs$\_$3s5z.
Stalkers are good at long-range attacks while Zealots are skilled in short-range attacks, and Stalkers move faster than Zealots.
Stalkers can win the game by simply attacking when the number of normal agents is sufficient, and they will fail if they just keep the same strategy under crashed scenarios. Figure~\ref{fig:obs2_1} illustrates that the Stalkers trained by QMIX actually only learn to attack continuously because this simple policy can achieve good performance under normal scenarios.
But if they can split into two groups, \emph{i.e.} some of them attract Zealots and do kitting ($i.e.$, attack and step back) repeatedly while others focus fire to eliminate the Stalkers and then attack remaining enemies together, they are likely to achieve better performance, as shown in Figure~\ref{fig:obs2_2}.
This case indicates that once a simple winning strategy exists, the learning algorithm has little incentive to explore other optimal strategies, leading to poor capability in the event of crashed agents.
The observation implies that increasing the challenge during training may drive the agents to learn better policies.

\begin{table*}[htbp]
    \scriptsize
	\centering
	\resizebox{\textwidth}{!}{
	\renewcommand{\arraystretch}{1.0}
	\begin{tabular}{|c|c|c|c|c|c|c|}%
		\hline
		\multirow{2}{*}{Method}  & \multicolumn{3}{|c|}{3s$\_$vs$\_$5z} & \multicolumn{3}{|c|}{3s5z$\_$vs$\_$3s5z} \\
		\cline{2-7}
		 &crash rate-0.01 & crash rate-0.05 & crash rate-0.10 & crash rate-0.01 & crash rate-0.05 & crash rate-0.10 \\
		\hline
		Baseline & 67.0 $\pm$ 15.5 & 61.9 $\pm$ 15.7 & 56.9 $\pm$ 15.3 & 85.3 $\pm$ 11.0 & 64.8 $\pm$ 8.3 & 43.6 $\pm$ 11.3 \\
		Fix-0.01 & 84.8 $\pm$ 11.8 & 74.7 $\pm$ 14.5 & 72.0 $\pm$ 13.8 & 86.9 $\pm$ 2.1 & 63.9 $\pm$ 1.7 & 45.8 $\pm$ 2.0 \\
		Fix-0.05 & 84.8 $\pm$ 7.5 & 78.0 $\pm$ 12.6 & 72.7 $\pm$ 9.6 & 86.3 $\pm$ 3.1 & 65.8 $\pm$ 2.9 & 46.9 $\pm$ 6.2 \\
		Fix-0.10 & 86.9 $\pm$ 8.0 & 81.1 $\pm$ 5.4 & 74.8 $\pm$ 8.2 & 83.6 $\pm$ 3.0 & 64.8 $\pm$ 7.0 & 48.1 $\pm$ 4.2 \\
		Curriculum & 84.4 $\pm$ 6.0 & 81.1 $\pm$ 8.1 & 74.7 $\pm$ 5.8 & 87.8 $\pm$ 2.5 & 66.1 $\pm$ 2.9 & 48.0 $\pm$ 1.6 \\
		Adaptive- & 82.5 $\pm$ 8.5 & 77.8 $\pm$ 8.4 & 71.6 $\pm$ 10.3 & 85.9 $\pm$ 4.5 & 65.2 $\pm$ 3.0 & 46.1 $\pm$ 1.9 \\
		\hline
		\textbf{Adaptive} & \textbf{88.6 $\pm$ 3.6} & \textbf{83.3 $\pm$ 6.5} & \textbf{79.2 $\pm$ 6.7}  & \textbf{88.0  $\pm$ 3.2} & \textbf{67.0  $\pm$ 2.4} & \textbf{51.7 $\pm$ 2.2} \\
		\hline
		\multirow{2}{*}{Method}  & \multicolumn{3}{|c|}{8m$\_$vs$\_$5z} & \multicolumn{3}{|c|}{8s$\_$vs$\_$3s5z} \\
		\cline{2-7}
		 &crash rate-0.01 & crash rate-0.05 & crash rate-0.10 & crash rate-0.01 & crash rate-0.05 & crash rate-0.10 \\
		\hline
		Baseline & 94.1 $\pm$ 2.3 & 82.3 $\pm$ 4.5 & 71.6 $\pm$ 2.9 & 88.6 $\pm$ 5.7 & 75.8 $\pm$ 7.0 & 68.6 $\pm$ 4.9 \\
		Fix-0.01 & 86.9 $\pm$ 5.4 & 79.8 $\pm$ 5.0 & 65.6 $\pm$ 4.4 & 87.5 $\pm$ 5.8 & 77.3 $\pm$ 6.6 & 62.0 $\pm$ 7.3 \\
		Fix-0.05 & 89.1 $\pm$ 2.4 & 84.2 $\pm$ 4.4 & 68.4 $\pm$ 6.4 & 91.1 $\pm$ 6.4 & 80.0 $\pm$ 7.2 & 66.7 $\pm$ 7.2 \\
		Fix-0.10 & 90.0 $\pm$ 5.0 & 83.9 $\pm$ 7.1 & 78.0 $\pm$ 4.8 & 88.9 $\pm$ 9.4 & 79.7 $\pm$ 11.3 & 70.0 $\pm$ 9.6 \\
		Curriculum & 94.1 $\pm$ 1.8 & 82.5 $\pm$ 2.8 & 72.3 $\pm$ 2.9 & 92.0 $\pm$ 2.9 & 79.8 $\pm$ 5.4 & 66.6 $\pm$ 5.8 \\
		Adaptive- & 91.3 $\pm$ 4.1 & 84.8 $\pm$ 2.1 & 78.6 $\pm$ 3.1 & 91.7 $\pm$ 3.1 & 80.0 $\pm$ 4.6 & 69.1 $\pm$ 6.0 \\
		\hline
		\textbf{Adaptive} & \textbf{94.2 $\pm$ 2.2} & \textbf{89.4 $\pm$ 2.4} & \textbf{81.1 $\pm$ 3.1}  & \textbf{93.9  $\pm$ 4.5} & \textbf{84.5 $\pm$ 10.0} & \textbf{71.3 $\pm$ 12.2} \\
		\hline
	\end{tabular}}
	\caption{The performance of the compared methods in terms of win rate (including mean and standard deviation) under different crash rates.
	Fix-i represents the variants of QMIX which indicates that the crashed rate is fixed to $i$ during training. Adaptive$-$ represents the results gained by adopting our adaptive method, but without the re-sampling strategy.}
	\label{tab:result}
\end{table*}

\subsubsection{Performance Evaluation and Discussion}


We evaluate the performance of the comparison methods by testing the win rate under different crash rates and the results are shown in Table~\ref{tab:result}. It can be observed that in standard maps, even just using a simple fixed crash rate strategy can help improve the performance. In contrast, in our designed maps, such approach works badly when the crash rate is low. We assume that it is because the maps we designed are relatively simple so that even original MARL algorithms can handle the scenarios with a low crash rate. In this case, fixing a low crash rate may instead introduce noise which affects the learning process. But in scenarios with a high crash rate, such method still has a positive effect. As for curriculum learning strategy, it tends to perform well in scenarios with a low crash rate. To sum up, these two straight methods can help the system be more robust in face of unexpected crash to some degrees but they all have some limits. On the contrary, our adaptive approach can help improve the performance in different maps and crash rates, which demonstrates the effectiveness and generalization of our approach. 

When compared with the baseline algorithm, our adaptive method tends to gain a greater margin when the crash rate increases, indicating the superiority of our adaptive strategy in dealing with unexpected crashes. This finding further implies the rationality of our adaptive strategy that allows the agents to learn how to handle the crash scenarios step by step. What's more, it can be observed that the performance achieved by our method with re-sampling has a consistent superiority, compared to the performance achieved without adopting this strategy. We think it can be attributed to the fact that without re-sampling strategy, there may be samples which contain more crashed agents, thus bringing more difficulty during training. This finding also proves the importance of adopting re-sampling strategy in our coach-assisted framework.

\subsubsection{Hyperparameter Analysis}\label{sec:hp}
In our adaptive framework, the performance threshold $\beta$ and the learning rate of crash rate $\rho$, which jointly decide the updating of the adaptive crash rate, are of vital importance to the performance of our method. In this part, we further analyze the influence of these two hyperparameters on the overall performance, with other parameters unchanged.

\begin{table*}[htbp]
	\centering
	\resizebox{\textwidth}{!}{
	\renewcommand{\arraystretch}{1.5}
	\begin{tabular}{|c|c|c|c|c|c|c|c|}%
		\hline
		 \multicolumn{4}{|c|}{$\rho=0.003$} & \multicolumn{4}{|c|}{$\beta = 0.65$} \\
		\cline{1-8}
		 --&crash rate-0.01 & crash rate-0.05 & crash rate-0.10 & -- & crash rate-0.01 & crash rate-0.05 & crash rate-0.10\\
		 \hline
	    $\beta = 0.6$ & 84.7 $\pm$ 9.0 & 80.5 $\pm$ 9.6 & 74.1 $\pm$ 8.3 & $\rho=0.001$ & 80.9 $\pm$ 9.4 & 75.6 $\pm$ 6.9 & 65.0 $\pm$ 13.0 \\
		\hline
		$\beta = 0.65$ & 88.6 $\pm$ 3.6 & 83.3 $\pm$ 6.5 & 79.2 $\pm$ 6.7 & $\rho=0.003$ & 88.6 $\pm$ 3.6 & 83.3 $\pm$ 6.5 & 79.2 $\pm$ 6.7 \\
		\hline
		$\beta = 0.7$ & 81.4 $\pm$ 7.8 & 77.7 $\pm$ 6.2 & 72.3 $\pm$ 7.1 & $\rho=0.005$ & 88.4 $\pm$ 4.5 & 83.6 $\pm$ 4.7 & 75.5 $\pm$ 8.6 \\
		 \hline
		 $\beta = 0.75$ & 79.5 $\pm$ 10.2 & 77.8 $\pm$ 8.5 & 67.0 $\pm$ 8.2 & $\rho=0.015$ & 78.4 $\pm$ 7.8 & 73.4 $\pm$ 8.5 & 68.9 $\pm$ 6.0 \\
		\hline
	\end{tabular}}
	\caption{The impact of performance threshold $\beta$ and learning rate $\rho$. The results show the win rate (including mean and standard deviation) under different settings across five different random seeds. The experiment takes QMIX as the base model on the 3s$\_$vs$\_$5z task after 2 million steps of training.}
	\label{tab:hp}
\end{table*}

Here, we take the map 3s$\_$vs$\_$5z as an example.
Table~\ref{tab:hp} reports the results of our method under different values of $\beta$ and $\rho$.
Given the same $\rho$, a large $\beta$ means that we require the agents to learn quite well under the current crash rate before exploring a harder scenario.
We can see that given $\rho=0.003$, the overall win rate first increases and then decreases as $\beta$ increases from 0.6 to 0.75, and the best performance is achieved when $\beta=0.65$.
Given the same $\beta$, we can see that the performance first improves and then degrades as the $\rho$ increases.
The reason may be that, if $\rho$ is too small, the crash rate $\alpha$ will be adjusted too slow, so that the agents cannot learn well within the limited steps.
If $\rho$ is too large, sharply increasing the crash rate may be too difficult for agents to learn coordination and the adjustment of the difficulty will be rough.
To sum up, the hyperparameters indeed have some effect on our framework, but our method can achieve a relatively stable performance if the hyperparameters are varied in a small range, which proves the robustness of our method.

\section{Conclusion}
\label{chap:conclusion}
Considering a common phenomenon that some agents may unexpectedly crash in real-world scenarios, this work is dedicated to a coach-assisted MARL framework that can close this sim-to-real gap.
Our method simulates different rates of random crashes during the training process with the help of ``coach'' so that the agents can master the skills to deal with crashes.
We conduct the experiments on grid-world and StarCraft II micromanagement tasks to show the necessity of considering crash during operation and test the effectiveness of our framework using three coaching strategies.
The results demonstrate the efficacy and generalization of our method under different crash rates.
In the future, we will further investigate the case in which the crashed agents may take other abnormal actions in addition to the random ones and other more efficient coaching strategies.

\bibliographystyle{unsrt}  
\bibliography{references}  







\end{document}